# Bayesian Random Fields: The Bethe-Laplace Approximation

**Max Welling**
Dept. of Computer Science
UC Irvine
Irvine CA 92697-3425
*welling@ics.uci.edu*

**Sridevi Parise**
Dept. of Computer Science
UC Irvine
Irvine CA 92697-3425
*sparise@ics.uci.edu*

## Abstract

While learning the maximum likelihood value of parameters of an undirected graphical model is hard, modelling the posterior distribution over parameters given data is harder. Yet, undirected models are ubiquitous in computer vision and text modelling (e.g. conditional random fields). But where Bayesian approaches for directed models have been very successful, a proper Bayesian treatment of undirected models in still in its infant stages. We propose a new method for approximating the posterior of the parameters given data based on the Laplace approximation. This approximation requires the computation of the covariance matrix over features which we compute using the linear response approximation based in turn on loopy belief propagation. We develop the theory for conditional and "unconditional" random fields with or without hidden variables. In the conditional setting we introduce a new variant of bagging suitable for structured domains. Here we run the loopy max-product algorithm on a "super-graph" composed of graphs for individual models sampled from the posterior and connected by constraints. Experiments on real world data validate the proposed methods.

## 1 INTRODUCTION

While much progress has been made in developing fully Bayesian methods for directed models (a.k.a. Bayesian networks), very little is known about Bayesian approaches for undirected models, (a.k.a. random fields). The main reason is computational; it is convenient to place conjugate priors on the parameters of a Bayesian network and marginalize them out analytically. Alternatively, one can resort to approximation schemes such as MCMC sampling and variational approximations.

The situation is much less favorable for undirected models. Computing a gradient of the log-likelihood is intractable due to the presence of the normalization constant that depends on the parameters of the model. So locating the ML solution is intractable for all but the simplest models having small treewidth. This fact implies that we cannot even marginalize over parameters approximately, by running a MCMC sampling method.

Random fields do however have an important role to play in modern AI. For instance in computer vision, interactions between pixels or patches have no causal directionality and are most naturally modelled by Markov random fields (MRF). Also, a popular language model for discriminative applications is the conditional random field (CRF) which has the advantage of being able to combine many interdependent features into a single probabilistic model. It is well known that in real world applications, CRFs suffer from overfitting. The most widely used heuristic to address that problem is to add a regularizer to the log-likelihood in the form of a log prior and search for the maximum a posteriori (MAP) solution.

Recently, a number of important "upgrades" for CRF models have been proposed. Firstly, for purely supervised learning, the probability of a label given an input is often less important than the actual prediction of that label. Hence, in a similar way that logistic regression and SVM relate to each other, one can formulate large margin classification methods for structured domains as an alternative for CRFs. These "max-margin Markov networks" ($M^3N$) (Taskar et al., 2003) can be kernelized and enjoy similar learning-theoretic guarantees as SVMs. Secondly, an approximation procedure based on the expectation propagation algorithm was proposed in (Qi et al., 2005a). The advantage of this approach is that instead of finding a single optimal set of model parameters, it approximately averages over all models weighted by their posterior probability. The basic assumption is that posterior distributions over parameters are Gaussian and that the extra coupling between labels induced by averaging over parameters can be well approximated by loopy belief propagation.

Apart from these developments in the context of CRFs, we know of only a few contributions that address the problem of Bayesian inference in Markov random fields. In (Murray & Ghahramani, 2004) a number of approximate MCMC samplers are proposed. Two of them were reported to be most successful: one based on Langevin sampling with approximate gradients given by contrastive divergence and one where the acceptance probability is approximated by replacing the log partition function with the Bethe free energy. Both these methods are very general, but inefficient. In (Green & Richardson, 2002) MCMC methods are explored for the Potts model based on the reversible jump formalism. To compute acceptance ratios for dimension-changing moves they need to estimate the partition function using a separate estimation procedure making it rather inefficient as well. In (Møller et al., 2006) a MCMC method is introduced that uses perfect samples to circumvent the calculation of the partition function altogether. This method is elegant but limited in its application due to the need to draw perfect samples.

In this paper we propose a new approach to approximate computations necessary for a full Bayesian treatment of random field models. Our approach is based on a second order series approximation of the log partition function around it's MAP solution, also known as the Laplace approximation. As the second order derivative of the log partition function w.r.t. its parameters is given by the full covariance matrix of the features, a further approximation is necessary. This is achieved through the linear response procedure reported in (Welling & Teh, 2004), where a message passing algorithm is proposed to approximate the covariance matrix of a random field. Since this linear response correction is in turn based on the Bethe approximation, we call the entire procedure the "Bethe-Laplace" approximation. The main assumptions are therefore that the posterior distribution over the parameters is well approximated by a normal distribution and moreover that belief propagation is a reasonable algorithm to approximate marginal distributions at the MAP solution.

## 2 RANDOM FIELD MODELS

Let $\mathbf{x}$ be a collection of discrete random variables. We define a random field model on these variables as,

$$p(\mathbf{x}|\boldsymbol{\lambda}) = \frac{1}{Z(\boldsymbol{\lambda})} \exp\left[\boldsymbol{\lambda}^T \mathbf{f}(\mathbf{x})\right] \quad (1)$$

where $\mathbf{f}(\mathbf{x}) = \{f_\alpha(\mathbf{x})\}$, $\alpha = 1..F$ is a vector of features, $\boldsymbol{\lambda} = \{\lambda_\alpha\}$, $\alpha = 1..F$ is a vector of parameters and $Z(\boldsymbol{\lambda})$ is the normalization constant or "partition function". The features define the potential functions of an undirected graphical model. The collection of variables inside the argument of a feature defines a clique in this graph. Although in principle features can be functions of all the variables in the problem, which would correspond to a fully connected graph, in practice we will assume that the graph is structured and the cliques generated by the features are small. For example, a Boltzman machine has features only on nodes and pairs of nodes in the graph[1]

We will take a Bayesian point of view and introduce a normal prior on the parameters $\boldsymbol{\lambda}$,

$$p(\boldsymbol{\lambda}) = \mathcal{N}(\boldsymbol{\lambda}; 0, \Lambda) \quad (2)$$

where $\mathcal{N}$ is a zero mean normal distribution with covariance $\Lambda$. Our goal is to find an approximation to the posterior distribution of the parameters given the data $\mathcal{D} = \{\mathbf{x}_n\}$, $n = 1..N$,

$$p(\boldsymbol{\lambda}|\mathcal{D}) \propto \mathcal{N}(\boldsymbol{\lambda}; 0, \Lambda) \prod_{n=1}^{N} \frac{1}{Z(\boldsymbol{\lambda})} \exp\left[\boldsymbol{\lambda}^T \mathbf{f}(\mathbf{x}_n)\right] \quad (3)$$

From this we recognize that the intractable partition function $Z(\boldsymbol{\lambda})$ represents the main obstacle for further progress. Indeed, the intractability of the partition function even prevents the execution of a standard MCMC sampler because the rejection rate depends on its evaluation. In the next section we will propose an approximate solution to this problem.

## 3 THE BETHE-LAPLACE APPROXIMATION

Our solution will involve two approximations: the Laplace approximation and the Bethe approximation.

The Laplace approximation replaces the complicated posterior with a normal distribution located at the maximum a posteriori value of the parameters, $\boldsymbol{\lambda}^{\text{MP}}$. The covariance is computed by expanding the log of the posterior up to second order. Noting that only the log partition function is a nonlinear function of $\boldsymbol{\lambda}$ we expand,

$$\log Z(\boldsymbol{\lambda}) \approx \quad (4)$$
$$\log Z(\boldsymbol{\lambda}^{\text{MP}}) + \boldsymbol{\kappa}^T(\boldsymbol{\lambda} - \boldsymbol{\lambda}^{\text{MP}}) + \frac{1}{2}(\boldsymbol{\lambda} - \boldsymbol{\lambda}^{\text{MP}})^T C(\boldsymbol{\lambda} - \boldsymbol{\lambda}^{\text{MP}})$$

with

$$\boldsymbol{\kappa} = \mathbb{E}[\mathbf{f}(\mathbf{x})]_{p(\mathbf{x})} \quad (5)$$
$$C = \mathbb{E}[\mathbf{f}(\mathbf{x})\mathbf{f}(\mathbf{x})^T]_{p(\mathbf{x})} - \mathbb{E}[\mathbf{f}(\mathbf{x})]_{p(\mathbf{x})}\mathbb{E}[\mathbf{f}(\mathbf{x})]_{p(\mathbf{x})}^T \quad (6)$$

where the averages are taken over $p(\mathbf{x}|\boldsymbol{\lambda}^{\text{MP}})$.

Assuming for a moment that we can compute or approximate $\boldsymbol{\lambda}^{\text{MP}}$ and $C$, and rewriting $\boldsymbol{\theta} = \boldsymbol{\lambda} - \boldsymbol{\lambda}^{\text{MP}}$, the final result of the Laplace approximation is,

$$p(\boldsymbol{\theta}|\mathcal{D}) \approx \mathcal{N}[\boldsymbol{\theta}; 0, \Sigma] \quad \text{with} \quad N\Sigma^{-1} = C + (N\Lambda)^{-1} \quad (7)$$

---
[1] Note that the maximal clique in a Boltzman machine can in fact correspond to the entire graph.

Note that this distribution has zero mean because $\boldsymbol{\lambda}^{\texttt{MP}}$ is the MAP solution. Also, given this observation, we do not need to compute $\boldsymbol{\kappa}$.

Unfortunately, the calculation of the quantities $\boldsymbol{\lambda}^{\texttt{MP}}$ and $C$ present intractable inference problems by themselves for graphical models with high treewidth. In the following we will assume that there is some method available to approximate $\boldsymbol{\lambda}^{\texttt{MP}}$. Candidates are pseudo-likelihood learning (Besag, 1977; Gidas, 1988), contrastive divergence (Hinton, 2002) or iterative scaling (Darroch & Ratcliff, 1972). The approximation of the covariance matrix $C$ is based on the linear response procedure and will be further discussed in the next section.

## 4 BP AND LINEAR RESPONSE

Computing the covariance matrix $C$ is intractable in general. Instead we use the linear response procedure explained in (Welling & Teh, 2004) to approximate it[2]. The main idea is to compute derivatives w.r.t. the parameters $\boldsymbol{\lambda}$ of the marginal probability distributions over the subsets of variables that are in the arguments of the features. These marginal distributions are themselves approximated using loopy belief propagation on factor graphs. There are two algorithms to convert these derivatives into a full covariance matrix, either by matrix inversion or by a message passing algorithm. The computational complexity of the propagation algorithm scales as $\mathcal{O}(F^2)$ per iteration with $F$ the number of features, while the matrix inversion naively scales as $\mathcal{O}(F^3)$. However, the matrix to be inverted is typically very sparse, so more efficient inversion ought to be possible as well.

In this paper we will restrict ourselves to binary variables and pairwise interactions which represents a particularly simple yet illustrative case. Due to space constraints, we leave the equations for the more general case with general discrete variables and arbitrary interactions for a subsequent publication. For binary variables and pairwise interactions we define the variables as $\boldsymbol{\lambda} = \{\theta_i, w_{ij}\}$ where $\theta_i$ is a parameter multiplying the node-feature $x_i$ and $w_{ij}$ the parameter multiplying the edge feature $x_i x_j$. Moreover, we'll define the following independent quantities $q_i = p(x_i = 1)$ and $\xi_{ij} = p(x_i = 1, x_j = 1)$. Note that all other quantities, e.g. $p(x_i = 1, x_j = 0)$ are functions of $\{q_i, \xi_{ij}\}$.

In (Welling & Teh, 2003) is was shown that in the Bethe approximation, these quantities are related to each other at the fixed points of BP,

$$w_{ij} = \log\left(\frac{\xi_{ij}(\xi_{ij}+1-q_i-q_j)}{(q_i-\xi_{ij})(q_j-\xi_{ij})}\right) \quad (8)$$

$$\theta_i = \log\left(\frac{(1-q_i)^{z_i-1}\prod_{j\in N(i)}(q_i-\xi_{ij})}{q_i^{z_i-1}\prod_{j\in N(i)}(\xi_{ij}+1-q_i-q_j)}\right) \quad (9)$$

---
[2]The procedure is exact for trees.

where $N(i)$ are neighboring nodes of node $i$ in the graph and $z_i = |N(i)|$ is the number of neighbors of node $i$. The first of these equations can be solved analytically for each $\xi_{ij}$ as a function of $\{q_i, w_{ij}, \theta_j\}$. (Welling & Teh, 2003).

To compute the covariance matrix we first compute,

$$C^{-1} = \begin{bmatrix} \frac{\partial \boldsymbol{\theta}}{\partial \mathbf{q}} & \frac{\partial \boldsymbol{\theta}}{\partial \boldsymbol{\xi}} \\ \frac{\partial \mathbf{w}}{\partial \mathbf{q}} & \frac{\partial \mathbf{w}}{\partial \boldsymbol{\xi}} \end{bmatrix} \quad (10)$$

and subsequently take its inverse[3]. In this expression, we set $\{q_i, \xi_{ij}\}$ to their values at the fixed point of BP. The derivatives required in 10 are unfortunately rather involved and are omitted due to space limitations. They will be published in a forthcoming journal publication while software will be put online http://www.ics.uci.edu/~welling/publications/publications.html.

The approximate estimate for $C$ as computed above is symmetric $C = C^T$ and positive semi-definite $C \succeq 0$ as one would expect for a covariance matrix (Welling & Teh, 2004). These properties are essential in ensuring that the posterior distribution that we wish to approximate is well defined and normalizable.

## 5 BAYESIAN CONDITIONAL RANDOM FIELDS

The class of conditional random field (CRF) models introduced by (Lafferty et al., 2001) defines a random field over labels $\mathbf{t}$ conditioned on input vectors $\mathbf{x}$ and as such does not need to model the details of the input distribution $p(\mathbf{x})$,

$$p(\mathbf{t}|\mathbf{x}, \boldsymbol{\lambda}) = \frac{1}{Z(\boldsymbol{\lambda}, \mathbf{x})} \exp\left[\boldsymbol{\lambda}^T \mathbf{f}(\mathbf{t}, \mathbf{x})\right] \quad (11)$$

Training of CRFs is only tractable if the undirected graphical model associated with the model has low treewidth. However, in practice overfitting is often an issue making regularization terms necessary. Hence, full Bayesian approaches seem to be a promising alternative in this domain. In fact, recently a Bayesian approach using an extension of the "expectation propagation" (EP) algorithm was proposed in (Qi et al., 2005a) and shown to significantly improve classification performance over the standard ML and MAP approaches.

To apply the proposed Laplace approximation to CRFs we expand exactly as in Eqn.4 replacing, $\log Z(\boldsymbol{\lambda}^{\texttt{MP}}) \rightarrow \log Z(\boldsymbol{\lambda}^{\texttt{MP}}, \mathbf{x})$, $\mathbf{f}(\mathbf{x}) \rightarrow \mathbf{f}(\mathbf{t}, \mathbf{x})$ and $p(\mathbf{x}) \rightarrow p(\mathbf{t}|\mathbf{x})$. Using this we can derive the approximate Gaussian posterior distribution over the parameters as being equal to 7 but with the following substitution: $C \rightarrow \frac{1}{N}\sum_{n=1}^{N} C_{\mathbf{x}_n}$ with

$$\begin{aligned}C_{\mathbf{x}_n} =& \mathbb{E}[\mathbf{f}(\mathbf{t}, \mathbf{x}_n)\mathbf{f}(\mathbf{t}, \mathbf{x}_n)^T]_{p(\mathbf{t}|\mathbf{x}_n)} - \\ & \mathbb{E}[\mathbf{f}(\mathbf{t}, \mathbf{x}_n)]_{p(\mathbf{t}|\mathbf{x}_n)} \mathbb{E}[\mathbf{f}(\mathbf{t}, \mathbf{x}_n)]^T_{p(\mathbf{t}|\mathbf{x}_n)}\end{aligned} \quad (12)$$

---
[3]With the boldface symbols $\mathbf{w}$ and $\boldsymbol{\xi}$ we mean the vector obtained by stacking the entries $w_{ij}$ or $\xi_{ij}$.

and where all averages are taken over distributions $p(\mathbf{t}|\mathbf{x}_n, \boldsymbol{\lambda}^{\text{MP}})$ at the MAP value $\boldsymbol{\lambda}^{\text{MP}}$ of the conditional log-likelihood $\sum_n \log p(\mathbf{t}_n|\mathbf{x}_n, \boldsymbol{\lambda})$.

We observe that we need to compute separate estimates for $C_{\mathbf{x}_n}$ for every data-vector, which is clearly computationally more intensive than the single estimate $C$ for "unconditional" models.

To predict the label for a new test case we are interested in computing or approximating,

$$p(\mathbf{t}_{n+1}|\mathbf{x}_{n+1}, \mathcal{D}) = \int d\boldsymbol{\lambda}\, p(\mathbf{t}_{n+1}|\mathbf{x}_{n+1}, \boldsymbol{\lambda})\, p(\boldsymbol{\lambda}|\mathcal{D}) \tag{13}$$

where the second term in the integral is given as a Gaussian distribution centered at $\boldsymbol{\lambda}^{\text{MP}}$ in the Bethe-Laplace approximation. We will assume that we are only interested in finding the optimal labels $\mathbf{t}_{n+1}$ for input $\mathbf{x}_{n+1}$ and not in the actual probabilities itself (e.g. as in classification). Hence, we can take the logarithm and ignore additive and multiplicative constants.

To approximate the integral we discuss two approaches in the following. Our first approach relies on expanding the log partition in the first term again as in Eqn.4, leading to the following objective function[4] for $\mathbf{t}_{n+1}$,

$$\begin{aligned}
\mathcal{G}(\mathbf{t}_{n+1}) =& \boldsymbol{\lambda}^{\text{MP}^T}\mathbf{f}(\mathbf{t}_{n+1}, \mathbf{x}_{n+1}) + \\
& \frac{1}{2}\mathbf{m}_{n+1}^T(C_{\mathbf{x}_{n+1}} + \Sigma^{-1})^{-1}\mathbf{m}_{n+1} \\
\text{with} \quad & \mathbf{m}_{n+1} = \mathbf{f}(\mathbf{t}_{n+1}, \mathbf{x}_{n+1}) - \boldsymbol{\kappa}_{n+1}
\end{aligned} \tag{15}$$

where $\Sigma$ is given by 7 and $\boldsymbol{\kappa}_{n+1} = \mathbb{E}[f(\mathbf{t}, \mathbf{x}_{n+1})]_{p(\mathbf{t}|\mathbf{x}_{n+1})}$.

The first term in expression 15 is exactly the expression we would have obtained from a standard MAP approach. Relative to MAP there is a positive quadratic correction term. We note that since $\Sigma = O(\frac{1}{N})$ this correction term vanishes at a rate $\frac{1}{N}$ as $N$ gets large.

Unfortunately, maximizing $\mathcal{G}(\mathbf{t})$ over the labels $\mathbf{t}$ is intractable, even if the undirected graph of the CRF is a tree. Take for example a simple case where features contain pairs of variables $f_{ij} \sim t_i t_j$. Then the quadratic correction term couples all such pairs resulting in interactions between quadruples of variables $t_i t_j t_k t_l$. Hence, in most cases the correction term will be hard to deal with computationally making additional approximations necessary. One possibility is to ignore a subset of the interaction terms, for instance only retaining pairwise interactions (note that new pairwise interactions are also generated from the correction term), or only retaining interactions between nodes close in the orig-

---

[4]In the derivation we used the following identity,

$$\frac{1}{2}x^T C x - a^T x = \frac{1}{2}(x - C^{-1}a)C(x - C^{-1}a) - \frac{1}{2}a^T C^{-1} a \tag{14}$$

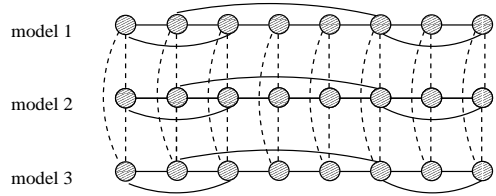

Figure 1: Super-graph based on three models sampled from the posterior. The max-product algorithm or any other approximate optimization algorithm should be run to obtain a good labelling solution. Solid edges are edges belonging to the CRF model while dashed edges are edges enforcing the constraint that associated variables in different models should have the same label.

inal graphical model. An algorithm such as "max-product" can then be used to maximize $\mathcal{G}(\mathbf{t})$.

An alternative approach is given by sampling a set $\mathcal{S}$ of parameter values from the posterior $p(\boldsymbol{\lambda}|\mathcal{D})$, which is Gaussian in the Bethe-Laplace approximation. The predictive distribution then becomes a mixture of random fields at these parameter values,

$$p(\mathbf{t}_{n+1}|\mathbf{x}_{n+1}, \mathcal{D}) = \frac{1}{|\mathcal{S}|}\sum_{s\in\mathcal{S}} p(\mathbf{t}_{n+1}|\mathbf{x}_{n+1}, \boldsymbol{\lambda}_s) \tag{16}$$

To make the maximization tractable we can interchange maximization and summation and take a majority vote over the individual maximizations. This is clearly suboptimal, as it does not take into account the confidence with which a particular model predicts a label. Hence, if each model separately is tractable (e.g. a tree), then we should weight the votes with their probabilities. The result is a bagging estimator where instead of bootstrap samples we take samples from the posterior. However, we propose a further improvement by constructing a "super-graph" as in figure 1, where associated labels in different models are connected by edges enforcing the constraint that they should have the same value. When we run the max-product algorithm on this super-graph, labels which are predicted with high probability in one model will transmit that confidence to other models using the inter-model connections. That way, a single very confident prediction can "convince" or "overrule" many weak predictions. This method depends on the max-product algorithm, or any other maximization method, to compute a good optimal solution for the labels on a loopy graph. But note, that the method will still work even if the individual models have cycles.

## 6 HIDDEN VARIABLES

Hidden variables are often a very powerful way to enhance the expressiveness of a probabilistic model. In the following we will discuss how to include these hidden variables in our formalism in both the unsupervised and the supervised setting.

For (unconditional) random fields with hidden variables $\mathbf{z}$ the probability distribution over the observed variables $\mathbf{x}$ looks like,

$$p(\mathbf{x}|\boldsymbol{\lambda}) = \int d\mathbf{z}\, \frac{\exp\left[\boldsymbol{\lambda}^T \mathbf{f}(\mathbf{x},\mathbf{z})\right]}{Z(\boldsymbol{\lambda})} = \frac{\mathcal{Z}(\mathbf{x},\boldsymbol{\lambda})}{Z(\boldsymbol{\lambda})} \quad (17)$$

where $\mathcal{Z}(\mathbf{x},\boldsymbol{\lambda})$ is the partition function for the conditional distribution $p(\mathbf{z}|\mathbf{x})$. Now expanding both $\log Z(\boldsymbol{\lambda})$ and $\log \mathcal{Z}(\mathbf{x},\boldsymbol{\lambda})$ up to second order around the MAP value, we obtain exactly the same expression for the posterior as in Eqn.7 but with the following substitution,

$$C \to C - \frac{1}{N}\sum_{n=1}^{N} C_{\mathbf{x}_n} \quad (18)$$

where $C$ is the covariance matrix w.r.t. $p(\mathbf{x},\mathbf{z})$ and $C_{\mathbf{x}_n}$ w.r.t.[5] $p(\mathbf{z}|\mathbf{x}_n)$.

For the fully observable case we have argued that for any value of $\boldsymbol{\lambda}$ (i.e. not necessarily the MAP value), $C$ is the second derivative of a globally convex function rendering it positive semi-definite (PSD). We have also argued that this result remains valid when we approximate $C$ using linear response (see (Welling & Teh, 2004) for a proof). In the case of hidden variables the situation is more complicated because the log-likelihood is no longer a globally convex function of $\boldsymbol{\lambda}$. Clearly, at the maximum likelihood solution the log-likelihood surface is locally convex, insuring that $C - \frac{1}{N}\sum_{n=1}^{N} C_{\mathbf{x}_n}$ is PSD at $\boldsymbol{\lambda} = \boldsymbol{\lambda}^{\text{MP}}$. This result is unfortunately no longer guaranteed when we replace $C$ and $\{C_{\mathbf{x}_n}\}$ with their linear response approximations.

To fix that, we first note that,

$$C^{\text{LR}} - \frac{1}{N}\sum_{n=1}^{N} C^{\text{LR}}_{\mathbf{x}_n} = \quad (19)$$
$$\frac{\partial^2}{\partial \lambda_i \partial \lambda_j}\left(G^{\text{BP}}[\boldsymbol{\lambda}, p(\mathbf{x},\mathbf{z})] - \frac{1}{N}\sum_{n=1}^{N} G^{\text{BP}}[\boldsymbol{\lambda}, p(\mathbf{z}|\mathbf{x}_n)]\right)$$

with $G^{\text{BP}}[\boldsymbol{\lambda}, p]$ the Bethe free energy (Yedidia et al., 2002). We will always assume that the Bethe free energy is minimized over the collection of marginal probability distributions [6] in its argument (subject to constraints) using belief propagation.

The difference between these two terms, called the contrastive Bethe free energy, is in general not convex in $\boldsymbol{\lambda}$. However, if we choose the parameters $\boldsymbol{\lambda}$ such that they maximize this approximate objective, then we regain local convexity at $\boldsymbol{\lambda}^{\text{MP,BP}}$ insuring the linear response approximation $C^{\text{LR}} - \frac{1}{N}\sum_{n=1}^{N} C^{\text{LR}}_{\mathbf{x}_n}$ to be PSD. In (Welling & Sutton, 2005) we have explored the problem of maximizing the contrastive Bethe free energy as a proxy to the ML objective and found that this is usually a reasonable procedure to follow.

---
[5] Beware of a potential confusion with the different definition of $C_{\mathbf{x}_n}$ for CRFs.

[6] For BP, this collection consists of node and pairwise marginals which are consistent with each other, but not necessarily consistent with a joint probability distribution over all variables.

For *conditional* random fields the situation is analogous. The probability distribution over the observable random variables $(\mathbf{t},\mathbf{x})$ is now given by,

$$p(\mathbf{t}|\mathbf{x},\boldsymbol{\lambda}) = \int d\mathbf{z}\, \frac{\exp\left[\boldsymbol{\lambda}^T \mathbf{f}(\mathbf{t},\mathbf{x},\mathbf{z})\right]}{Z(\mathbf{x},\boldsymbol{\lambda})} = \frac{\mathcal{Z}(\mathbf{t},\mathbf{x},\boldsymbol{\lambda})}{Z(\mathbf{x},\boldsymbol{\lambda})} \quad (20)$$

where $\mathcal{Z}(\mathbf{t},\mathbf{x},\boldsymbol{\lambda})$ is the partition function for the conditional distribution $p(\mathbf{z}|\mathbf{t},\mathbf{x})$. Following the now familiar procedure of expanding both partition functions we arrive again at Eqn.7 with the following substitution,

$$C \to \frac{1}{N}\sum_{n=1}^{N}(C_{\mathbf{x}_n} - C_{\mathbf{t}_n,\mathbf{x}_n}) \quad (21)$$

where $C_{\mathbf{x}_n}$ is the covariance matrix w.r.t. $p(\mathbf{t},\mathbf{z}|\mathbf{x}_n)$ and $C_{\mathbf{t}_n,\mathbf{x}_n}$ the covariance matrix w.r.t. $p(\mathbf{z}|\mathbf{t}_n,\mathbf{x}_n)$.

Again we notice that we can only guarantee that this covariance matrix is PSD in the linear response approximation if we estimate the parameters $\boldsymbol{\lambda}$ by minimizing the conditional contrastive Bethe free energy,

$$\frac{1}{N}\sum_{n=1}^{N}\left(G^{\text{BP}}[\boldsymbol{\lambda}, p(\mathbf{t},\mathbf{z}|\mathbf{x}_n)] - G^{\text{BP}}[\boldsymbol{\lambda}, p(\mathbf{z}|\mathbf{t}_n,\mathbf{x}_n)]\right) \quad (22)$$

where we refer to (Welling & Sutton, 2005) for details on how to optimize this objective.

## 7 EXPERIMENTS

To assess the accuracy of the estimated posterior distributions we decided to compare sample sets. We used a scoring function between two sample sets based on the Cramer-Von Mises test statistic. For finite univariate sample sets $S_1$ and $S_2$ of size $N_1$ and $N_2$ respectively, the score is defined as:

$$\text{CVM}(S_1, S_2) = \sum_{i=1}^{N_1+N_2}(F_{S_1}(X_i) - F_{S_2}(X_i))^2 \quad (23)$$

where $\{X_i\}$ are all the samples in $S1 \cup S2$ and $F_{S_1}$ and $F_{S_2}$ are the empirical CDFs based on $S_1$ and $S_2$. For the overall multivariate score we just take the sum of scores over all dimensions. Note that this measure has the desirable property that it does not depend on an arbitrary bin-size or smoothing parameter.

The following four methods were compared:
-**BL-MP:** The Bethe-Laplace approximation where we provide the exact MAP value as estimated using the junction tree algorithm.

-**BL-BP:** The Bethe-Laplace approximation where we estimate the MAP value by approximating the gradient using loopy BP. This was implemented by first ignoring the prior and using pseudo moment matching (M.J. Wainwright & Willsky, 2003) to approximate the ML estimate. Initializing on that value we then perform gradient descend on the penalized log-likelihood (including the prior) and approximating necessary averages in the gradient using loopy BP.

-**LV-CD:** The brief approximate Langevin sampler proposed in (Murray & Ghahramani, 2004) where we approximate the required gradient using contrastive divergence.

-**MC-BP:** A standard MCMC sampler with symmetric Gaussian proposals where the Metropolis-Hastings accept/reject rule is implemented by replacing the partition function with the Bethe free energy (Murray & Ghahramani, 2004).

To have access to ground truth we first looked at $5 \times 5$ square grid models with binary variables ($\pm 1$). Fair samples from the posterior of this model were drawn by using a junction tree to compute the exact gradient of the penalized log-likelihood (including a prior term) which was used in the "hybrid Monte Carlo" (HMC) algorithm to draw the samples.

In the first experiment we randomly draw 5 parameter sets (weights and biases as in Boltzman machines) from a zero mean normal distribution with covariance $0.25 \times I$. For all methods, 10 sample sets of 10K samples were obtained for each model. For the two methods based on the Bethe-Laplace approximation we draw 10 sets of 10K independent samples from the normal approximation of the posterior. Both sampling based methods were restarted 5 times and sub-sampled at intervals approximately equal to the auto-correlation time of the respective MCMC chains, also producing a total of 10 sets of 10K almost independent samples. We estimated convergence of the Markov chains with the "multivariate potential scale reduction factor" (MPSRF) using a threshold of $1.1$. The sample sets were subsequently used in the CVM scoring function to compare their deviation from ground truth. To compute a final score we average the individual scores over the 10 sample sets and over the 5 models. Data-cases were obtained by drawing perfect samples using the "forward prediction backward sampling" algorithm on the junction tree. Scores as a function of the number of data-cases are shown in figure 2. We note the increase in score for **BL-MP** at $N = 5000$ which was caused by non-Gaussian marginal distributions, see figure 3. Our implementations of **LV-CD** and **MC-BP** took between $6 - 16$ hours for each $N$ - exact times varying based on convergence times of the Markov chains and $N$, while those of **BL-MP** and **BL-BP** took less that 2 minutes (all in MATLAB). Note however that these times are very implementation dependent, especially for **LV-CD** and **MC-BP** since tricks to speed up mixing and convergence of the chains could probably bring down running times.

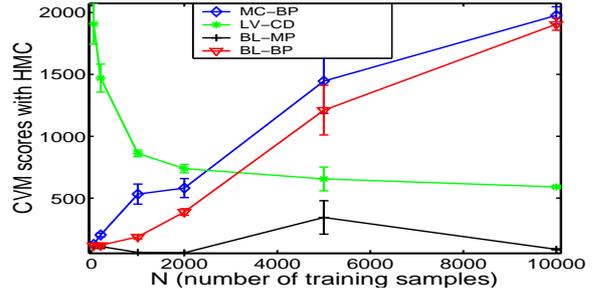

Figure 2: CVM score as a function of the number of data-cases for the four methods described in the main text.

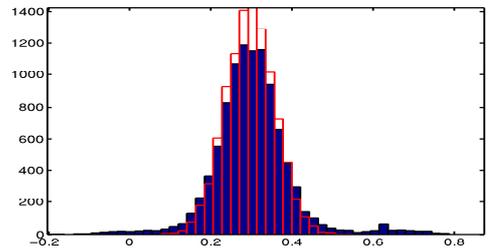

Figure 3: Histograms for **BL-MP** (transparent) and HMC (filled) at $N = 5000$ in figure 2. Note the non-Gaussian tails which cause the increase in score.

In a second experiment we study the dependence on the interaction strength by sampling weights from $\mathcal{U}([-(d+\varepsilon), -d] \cup [d, d+\varepsilon])$ and varying $d$. We have used $\varepsilon = 0.1/4$ and $d = [0.1, 0.5, 1, 1.5, 2]/4$. Biases are all set to either 0 in one experiment or 1 in another experiment, corresponding to figures 4 and 5. Note that for zero bias loopy BP failed to converge around $d = 1.5$. Averages and error bars in this experiment are computed over 10 sample sets but for a single instantiation of parameters. Running times were similar to previous experiment (except where loopy BP failed to converge).

From these experiments we conclude that the Bethe-Laplace approximation is very accurate if one can find an accurate approximation to the MAP value of the parameters. Using loopy BP to find an approximate MAP value works well for small sample sizes and small edge-strength. It is interesting to see that the score increases when we increase the number of data-cases. This effect is caused by the fact that the inherent bias in the BP estimate of the MAP value becomes more exposed as the variance of the posterior decreases with growing $N$. We see that **LV-CD** is less accurate in the "easy domain" but more robust to the increasing interaction strength and sample set size than BP

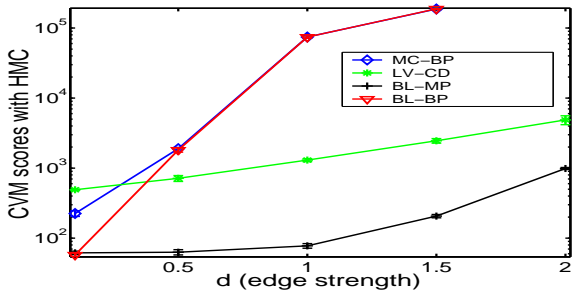

Figure 4: CVM score as a function of the interaction strength with all biases set to 0.

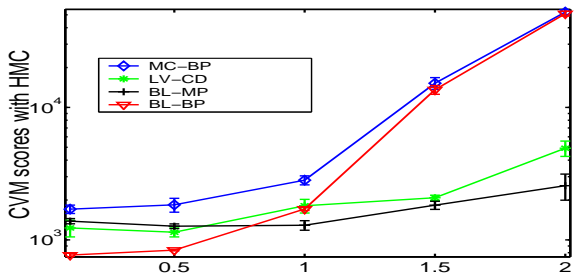

Figure 5: CVM score as a function of the interaction strength with all biases set to 1.

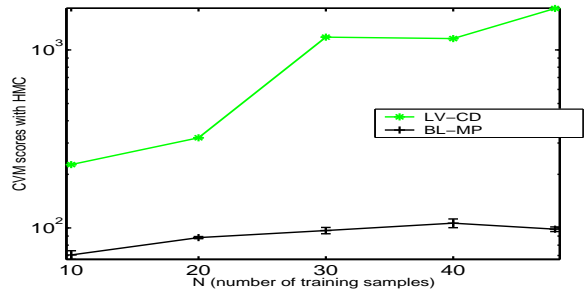

Figure 6: CVM scores for **BL-MP** and **LV-CD** for the CRF model. Note that we left out BP related methods because BP is exact on a chains which implies that **MC-BP** is equivalent to the MCMC sampler we use as ground truth and **BL-BP** is equivalent to **BL-MP**.

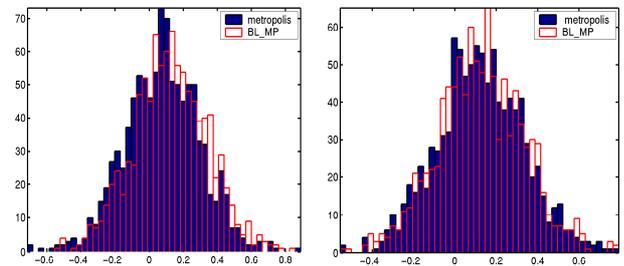

Figure 7: Worst (left) and best (right) scoring fits of posterior marginal distributions for CRF model. We compare **BL-MP** (transparent) with MCMC sampling (filled) on $N = 10$ files.

based methods.

To see if the proposed method also works well on real world data we implemented a CRF on the "newsgroup FAQ dataset"[7] (McCallum & Pereira, 2000). This dataset contains 48 files where each line can be either a header, a question or an answer. We binarized this problem by only retaining the question/answer lines. All files are truncated at 100 lines. For each line we extract 24 binary features $g^a(x) = 0/1$, $a = 1,..,24$ as provided by (McCallum & Pereira, 2000). These are used to define state and transition features using: $f_i^a(t_i, x_i) = t_i g^a(x_i)$ and $f_i^a(t_i, t_{i+1}, x_i) = t_i t_{i+1} g^a(x_i)$ where $i$ denotes the line in a document and $a$ indexes the 24 features. After adding a prior, we then run a standard gradient optimizer (using forward-backward to do the required inference) to find the MAP value. Classification error using this MAP value is about $1.5\%$ on average. Parameters in this model are tied in the sense that they only depend on the feature index $a$ giving a total of 48 parameters (24 state and 24 transition features). To deal with that we first extend all files to have exactly 100 lines by assuming features always take on the value 0 for empty lines (this does not affect the results). Furthermore, we first assume that the weights and biases are untied and compute a full $(99 + 100) \times (99 + 100)$ dimensional covariance matrix which we subsequently project down into a $48 \times 48$ dimensional submatrix. A standard Metropolis-Hastings MCMC sampler is used to obtain samples from the posterior where we use subsets of size $N = [10, 20, 30, 40, 48]$ as training data. This is then compared against samples obtained using **BL-MP** and **LV-CD**. The brief Langevin method uses 1 step of a Gibbs sampler (updating all variables in the CRF chain once) to compute the gradients. In addition, figures 7 and 8 show the worst and best fit according to the CVM score for $N = 10, 48$. Running times for each $N$ were around 5 hours and 2 minutes for **LV-CD** and **BL-MP** respectively.

---

[7] Downloaded from: http://www.cs.umass.edu/~mccallum/data/faqdata/

## 8 DISCUSSION

Our contribution in this paper has been to propose and test a new technique to approximate the posterior distribution over parameters in random field models. Since this issue has received so little attention in the literature, we feel new ideas to approach this very difficult problem are important. We have shown that our method can work very well, in particular when we have a good estimate of the MAP value of the parameters. It is noteworthy that our method was many orders of magnitude faster than the approximate sampling

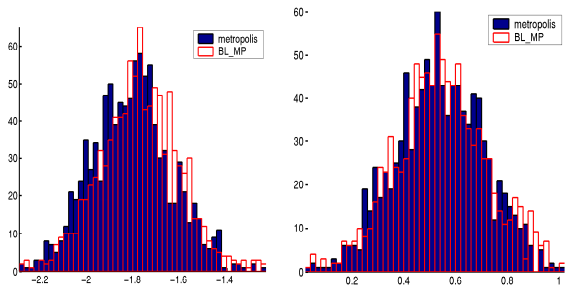

Figure 8: Same as figure 7 for $N = 48$ files.

schemes that we tested against.

The computational complexity of our Bethe-Laplace approximation naively scales as $\mathcal{O}(F^3)$ (with $F$ the number of features) because we need to compute a matrix inverse of a $F \times F$ matrix. However, $C$ and its inverse can be computed in $\mathcal{O}(F^2 S)$ using the linear response algorithm where $S$ represents the maximal number of variables in the argument of a feature. The underlying reason seems to be the sparseness of $C^{-1}$ due to locality of the features. Depending on the prior we therefore expect that better scaling for the entire BL procedure is possible.

There are a number of interesting extensions to consider. Firstly, the Bethe approximation can be replaced with higher order Kikuchi approximations. Linear response approximations have been developed for this case as well (Tanaka, 2003) and could be integrated into a similar "Kikuchi-Laplace" approximation. Secondly, the Bayesian setup allows in principle to prune features that are uninformative. This pruning could be implemented by using an ARD type prior. Success with such an approach has been reported for the EP approximation in the context of CRF models (Qi et al., 2005b).

An open question is whether the approach can also be useful for model selection. The evidence still depends on the partition function at its MAP value which makes its evaluation intractable. Replacing it with the Bethe or Kikuchi free energy is an option that we haven't yet explored.

**Acknowledgments**

This work was supported by a NSF-Career grant, IIS 0447903. P. Baldi and R. Lathrop are gratefully acknowledged for granting access to their clusters. We thank Yee Whye Teh for interesting discussions.